\documentclass{article}

\usepackage[preprint]{neurips_2026}

\usepackage[utf8]{inputenc}
\usepackage[T1]{fontenc}
\usepackage{hyperref}
\usepackage{url}
\usepackage{booktabs}
\usepackage{amsfonts}
\usepackage{amsmath}
\usepackage{graphicx}
\usepackage{microtype}
\usepackage{xcolor}

\hypersetup{
  colorlinks=true,
  linkcolor=blue,
  citecolor=blue,
  urlcolor=blue
}

\newcommand{\method}{EMG-GPT}

\title{EMG-GPT: Predictive Pretraining on Residual-Quantized EMG Tokens for Hand Pose Estimation}

\author{Ettore Magni \textsuperscript{*} \\
Imperial College London \\
\And
Rolandos Alexandros Potamias \\
Imperial College London \\
\AND
Stefanos Zafeiriou \\
Imperial College London \\
\And
Konstantinos Barmpas \\
Imperial College London \\
}

\begin{document}

\maketitle

\renewcommand{\thefootnote}{\fnsymbol{footnote}} 
\footnotetext[1]{Conducted during an internship at Imperial.}

\begin{abstract}
  Surface electromyography (sEMG) is a low-power, cost-effective biosignal for hand-pose estimation and gesture classification. In this work, we examine whether self-supervised pretraining on sEMG can yield transferable representations for continuous hand-pose estimation. We introduce EMG-GPT, a causal transformer-based model that operates on discrete sEMG representations from a frozen residual vector quantization (RVQ) tokenizer and learns temporal dynamics through depth-autoregressive future-code prediction. The model combines within-frame integration with causal temporal modeling while preserving the geometry of the pretrained codebook. EMG-GPT shows competitive results in both Regression and Tracking tasks, supporting EMG-only pretraining as a viable approach for learning transferable sEMG representations.
\end{abstract}

\section{Introduction}
\label{sec:introduction}

Hand pose estimation and gesture classification on wearable devices have attracted increasing interest due to their potential in a wide range of applications, particularly in human--computer interaction ~\citep{s21093035}. Although camera-based approaches can achieve high accuracy ~\citep{pavlakos2023reconstructinghands3dtransformers,potamias2025wilorendtoend3dhand,si2026anyhand}, they are susceptible to occlusions, which can compromise the stability and reliability of vision-based estimations and predictions. These limitations have motivated the exploration of alternative sensing modalities ~\citep{Tchantchane}. Surface electromyography (sEMG) measures the electrical activity generated during muscle activation, providing a wearable and non-invasive signal for estimating hand movements.

Unlike gesture classification, continuous pose estimation must recover an evolving articulated state. Existing systems learn supervised mappings from
sEMG to joint trajectories~\citep{quivira2018semgpose,liu2021neuropose,
simpetru2022dof}. The emg2pose benchmark ~\citep{salter2024emg2pose} supports this task at unusual scale, with 370 hours of paired sEMG and motion capture from 193 participants and
generalization tests across users, sensor placements and activity stages. Such supervised results establish task-specific
decoding performance, but do not by themselves show whether unlabelled EMG
supports reusable predictive representations. We therefore study whether
temporal pretraining with EMG-derived targets can produce states that transfer
to motor decoding without exposing the representation learner to kinematic
labels. Downstream evaluation then tests what these states retain when
pose-specific supervision is introduced.

Many biosignal pretraining approaches learn from masked or reconstructed signal
segments~\citep{yang2023biot,cui2024neurogpt,jiang2024labram} while autoregressive pretraining has also been explored for electroencephalography (EEG)~\citep{yue2025braingpt}.
NeuroRVQ~\citep{barmpas2025neurorvq} provides a complementary discrete interface: it decomposes biosignals
into frequency-specific representations, encodes them with an ordered residual
vector quantization (RVQ) hierarchy and explores masked-token modeling over the resulting codes. Its output representations are structured over electrodes, frequency branches and residual depth rather than drawn from one flat vocabulary, making preservation of that structure necessary and part of the modeling problem.

Building on the NeuroRVQ work, we introduce \method{}, a causal predictor of structured EMG representation tokens. By keeping
the pre-trained NeuroRVQ-EMG tokenizer and codebooks fixed, we study the temporal representation learned above this interface. Rather than flattening every electrode--branch--depth code into the temporal
sequence, \method{} forms one temporal state per EMG frame, keeps sequence length independent of the retained RVQ depth $K$ and predicts future-frame codes in RVQ depth order against the corresponding frozen codebooks. Keeping the discrete interface fixed separates temporal pretraining from tokenizer adaptation.

However, token predictability is not equivalent to kinematic utility. A model may reduce predictive cross-entropy by exploiting token regularities unrelated to hand motion.
We test movement relevance by transferring the learned states to
the emg2pose Regression and Tracking tasks under held-out generalization
conditions. Our contributions are: 
\begin{enumerate}
    \item \method{}, a frame-level causal future-prediction formulation based on NeuroRVQ-EMG tokens whose temporal sequence length is independent of retained RVQ depth $K$
    \item a depth-autoregressive prediction mechanism that conditions finer RVQ levels on coarser levels and scores them against their corresponding frozen codebooks
    \item a controlled transfer study across retained depth and pretraining progress on continuous hand-pose Regression and Tracking tasks\end{enumerate}


\section{Background}
\label{sec:background}

\subsection{Frozen NeuroRVQ EMG representations}

NeuroRVQ maps a multi-channel (C channels) EMG time window to $B$ frequency branches and applies a
branch-specific RVQ codebook stack (K layers) at every electrode~\citep{barmpas2025neurorvq}. Let
$\mathbf{E}_{b,k}\in\mathbb{R}^{M\times d_e}$ be the codebook for branch $b$
and depth $k$, where $M$ denotes the number of tokens  and $d_e$ is the token dimension. The integer-valued code index
$q_{t,c,b,k}\in\{0,\ldots,M-1\}$ selects a row of $\mathbf{E}_{b,k}$ at frame
$t$, electrode $c$, branch $b$, and depth $k$, and has no geometric or ordinal
meaning. For retained depth $K$, the structured frame array is:
\begin{equation}
  \mathbf{Q}_t=(q_{t,c,b,k})_{c,b,k}
  \in\{0,\ldots,M-1\}^{C\times B\times K}
  \label{eq:bg-token-array}
\end{equation} 

The reconstructed tokenizer vector at each
electrode--branch position is:

\begin{equation}
    \widehat{\mathbf{z}}^{(K)}_{t,c,b}
=\sum_{k=1}^{K}\mathbf{E}_{b,k}[q_{t,c,b,k}]
\end{equation} 

We refer the reader to Appendix~\ref{app:rvq} for a detailed analysis of the
NeuroRVQ-EMG token representation.


\subsection{Structured causal prediction}

Equation~\eqref{eq:bg-token-array} is not a one-dimensional language sequence:
each time step contains electrode, branch, and residual-depth axes. A temporal
predictor must therefore model causal dependencies across frames while
accounting for the ordered residual structure within the frame being predicted.
This creates a representation-learning problem with two distinct structures:
\emph{(i)} time-ordered context and \emph{(ii)} a branch-specific coarse-to-fine code hierarchy. 

\newpage
\section{Method}
\label{sec:method}

\method{} uses the produced NeuroRVQ tokens and its codebook geometry to represent each EMG frame. \textbf{A bidirectional intra-frame encoder}
integrates simultaneous electrode--frequency features into one frame state,
while \textbf{a causal transformer} models their temporal evolution. EMG-only
pretraining predicts future codes coarse-to-fine across RVQ depth. 
Tensor-level analysis is described in Appendix~\ref{app:emg-gpt-interface}.

\begin{figure*}[!h]
  \centering
  \includegraphics[width=\textwidth]{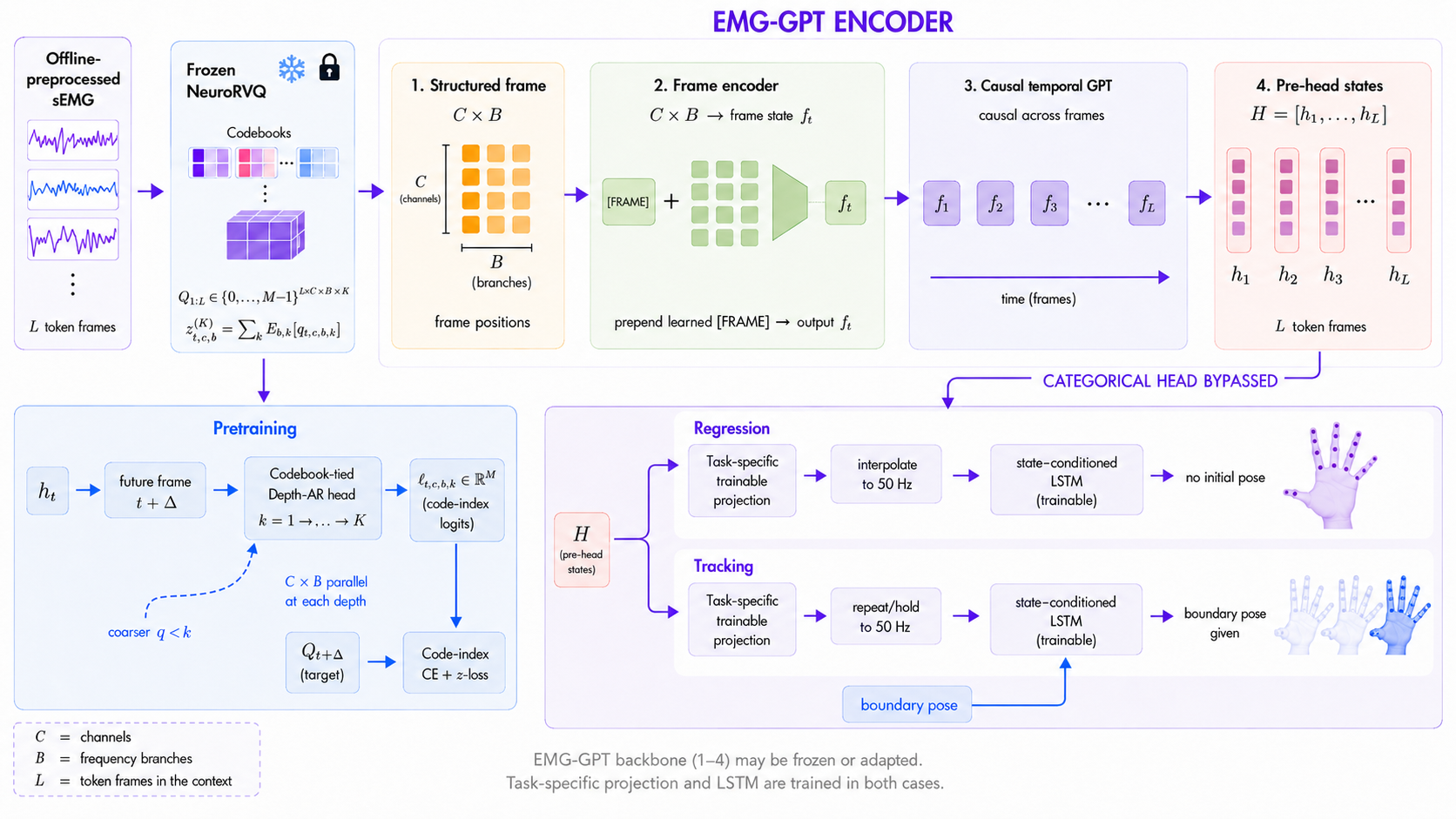}
  \caption{Overview of \method{}. sEMG patches are mapped by frozen NeuroRVQ to structured time--electrode--branch--depth codes. Frozen vector sums are integrated within each frame and modeled causally across frames. The pre-head
  state either predicts a future frame coarse-to-fine with code-index cross-entropy,
  or bypasses the categorical head and transfers to parallel Regression and
  Tracking decoders under frozen or full-backbone adaptation. Regression
  interpolates features to the $50$-Hz pose grid without an initial pose, whereas
  Tracking repeats features and receives the boundary pose. Only token-frame
  modeling is causal.}
  \label{fig:method-overview}
\end{figure*}

\subsection{Structured RVQ Frame Encoding}

From $\mathbf{Q}_t$ in Eq.~\eqref{eq:bg-token-array}, we retain any prefix
$1\leq K\leq K_{\max}$, where $K_{\max}$ is the tokenizer's full RVQ depth.
Its frozen RVQ reconstruction is:
\begin{align}
  \widehat{\mathbf{z}}^{(K)}_{t,c,b}
  &=\sum_{k=1}^{K}\mathbf{E}_{b,k}[q_{t,c,b,k}],
  \label{eq:method-rvq-sum}\\
  \mathbf{x}_{t,c,b}
  &=\mathbf{W}_{\mathrm{in}}\,\operatorname{LN}\!\left(
    \widehat{\mathbf{z}}^{(K)}_{t,c,b}\right)
    +\mathbf{e}^{\mathrm{ch}}_c+\mathbf{e}^{\mathrm{br}}_b
  \label{eq:method-frame-input}
\end{align}
Here $d_{\mathrm{model}}$ is the model hidden dimension,
$\operatorname{LN}$ is a LayerNorm layer over the reconstructed $\widehat{\mathbf{z}}^{(K)}_{t,c,b}$ vector and
$\mathbf{W}_{\mathrm{in}}\in\mathbb{R}^{d_{\mathrm{model}}\times d_e}$ is a
learned bias-free projection from the $d_e$ dimension to the $d_{\mathrm{model}}$ dimension.
Finally, $\mathbf{e}^{\mathrm{ch}}_c$ and $\mathbf{e}^{\mathrm{br}}_b$ are learned channel
and frequency-branch embeddings. Residual vector
summation follows the ordered RVQ construction~\citep{zeghidour2022soundstream,
barmpas2025neurorvq}, whereas the normalization, projection, and structural embeddings
define our model input.

We define the ordered frame sequence $\mathbf{X}_t$ as a channel-major and branch-minor sequence:
\begin{equation}
  \mathbf{X}_t=[\mathbf{x}_{t,1,1},\ldots,\mathbf{x}_{t,1,B},
  \mathbf{x}_{t,2,1},\ldots,\mathbf{x}_{t,C,B}]
  \in\mathbb{R}^{CB\times d_{\mathrm{model}}}
  \label{eq:method-ordered-frame}
\end{equation}
Given a learned $\mathrm{[FRAME]}$ token and applying the intra-frame transformer gives:
\begin{equation}
  \mathbf{f}_t=\left[\operatorname{LN}_{\mathrm{fr}}
  F_{\mathrm{fr}}([\mathbf{e}^{\mathrm{[FRAME]}};\mathbf{X}_t])
  \right]_{\mathrm{[FRAME]}}
  \label{eq:method-frame-state}
\end{equation}
$F_{\mathrm{fr}}$ denotes the bidirectional intra-frame Transformer,
$\operatorname{LN}_{\mathrm{fr}}$ its final LayerNorm,
$\mathbf{e}^{\mathrm{[FRAME]}}$ the learned frame-summary token and
$[\cdot]_{\mathrm{[FRAME]}}$ selects its output, yielding
$\mathbf{f}_t\in\mathbb{R}^{d_{\mathrm{model}}}$ for frame $t$.

The $CB$ positions are simultaneous electrode--frequency features, making the intra-frame attention  \textit{bidirectional}.
Since RVQ depth has already been combined before this encoder, increasing $K$ enriches each frame and adds prediction targets without increasing
the $CB$ positions or the temporal sequence length.

\subsection{Hierarchical Predictive Pretraining}

For a context of $L$ token frames, learned absolute temporal embeddings are
added to the frame states before a decoder-only transformer:
\begin{equation}
  \mathbf{h}_{1:L}
  =\operatorname{LN}_{\mathrm{out}}F_{\mathrm{GPT}}\!\left(
  \operatorname{Drop}([\mathbf{f}_t+\mathbf{e}^{\mathrm{time}}_t]_{t=1}^{L})
  \right)
  \label{eq:method-temporal}
\end{equation}
where $\mathbf{e}^{\mathrm{time}}_t$, $\operatorname{Drop}$, $F_{\mathrm{GPT}}$
and $\operatorname{LN}_{\mathrm{out}}$ denote the temporal embedding, dropout,
decoder-only temporal Transformer, and final LayerNorm, respectively, and
$\mathbf{h}_t\in\mathbb{R}^{d_{\mathrm{model}}}$ is the resulting frame state.
The causal mask~\citep{vaswani2017attention} restricts $\mathbf{h}_t$ to frame
states through $t$. Training pairs are denoted with $\mathbf{Q}_{1:L}$ and
$\mathbf{Q}_{1+\Delta:L+\Delta}$, where $\Delta$ is an offset in token frames.
If adjacent frames are separated by $\delta t$, the forecast horizon is
$\tau_{\mathrm{forecast}}=\Delta\delta t$. We use a 25-Hz configuration $(5,40\,\mathrm{ms})$, predicting 200~ms ahead. We therefore describe
the temporal objective generically as causal future-frame prediction. 
Prediction follows RVQ depth because each finer code describes a residual left
by coarser codes~\citep{lee2022rqtransformer}. For target depth $k$, define the
reconstruction from realized coarser codes:
\begin{equation}
  \widehat{\mathbf{z}}^{(<k)}_{t+\Delta,c,b}
  =\sum_{r=1}^{k-1}\mathbf{E}_{b,r}[q_{t+\Delta,c,b,r}],
  \qquad \widehat{\mathbf{z}}^{(<1)}=\mathbf{0}
  \label{eq:method-depth-context}
\end{equation}
Given $\mathbf{Q}_{\leq t}$ as the context, $\mathbf{Q}_{t+\Delta,:,:,<k}$ the
earlier target depths and $p_\theta$ as the model distribution, the factorization is defined as:
\begin{equation}
  p_\theta(\mathbf{Q}_{t+\Delta}\mid\mathbf{Q}_{\leq t})
  =\prod_{k=1}^{K}\prod_{c=1}^{C}\prod_{b=1}^{B}
  p_\theta\!\left(q_{t+\Delta,c,b,k}\mid\mathbf{Q}_{\leq t},
  \mathbf{Q}_{t+\Delta,:,:,<k}\right)
  \label{eq:method-depth-factorization}
\end{equation}
Bidirectional
self-attention exchanges hidden information among same-depth queries before
cross-attention to $\mathbf{h}_t$. Thus, the model is: \emph{(i)} causal across time, \emph{(ii)}  autoregressive across RVQ depth and \emph{(iii)} 
conditionally factorized across channel--branch targets within a depth. Training uses ground-truth coarser codes while generation uses previously predicted ones.

Each query is projected into codebook space and scored against its matching
frozen branch--depth codebook by learned scaled \textit{cosine similarity}: code indices provide
categorical targets, while codebook vectors are the output prototypes. With logits
$\boldsymbol{\ell}_{n,t,c,b,k}\in\mathbb{R}^{M}$, the objective for $N$ complete
windows is:
\begin{equation}
  \begin{aligned}
  \mathcal{L}_{\mathrm{CE},k}
  &= -\frac{1}{NLCB}\sum_{n,t,c,b}
  \log\operatorname{softmax}(\boldsymbol{\ell}_{n,t,c,b,k})_
  {q_{n,t+\Delta,c,b,k}}, 
  \\
  \mathcal{L}_{\mathrm{pre}}
  &= \frac{1}{K} \sum_{k=1}^{K}
  \left(\mathcal{L}_{\mathrm{CE},k}+\lambda_z\mathcal{L}_{z-loss}\right)
  \end{aligned}
  \label{eq:method-pretraining-loss}
\end{equation}

We use $\lambda_z=10^{-4}$ for z-loss regularization term. 
We report the clean predictive metric
$\mathcal{L}_{\mathrm{CE}}=\frac{1}{K} \sum_k\mathcal{L}_{\mathrm{CE},k}$, while
optimization uses $\mathcal{L}_{\mathrm{pre}}$. Complete shifted windows require
no padding mask. No gradient reaches NeuroRVQ
or its codebooks. Code-prediction loss trains the backbone, but pose transfer
tests the pre-head states and bypasses this categorical objective. 

\subsection{Transfer to Hand Pose}

Downstream computation uses the
$\mathbf{H}=[\mathbf{h}_1,\ldots,\mathbf{h}_L]
\in\mathbb{R}^{N\times L\times d_{\mathrm{model}}}$ as described in Section \ref{sec:method}. A learned feature projection changes representation width, while temporal
rate alignment is a separate operation. \emph{Regression} linearly
interpolates projected features to the 50-Hz pose grid, receives no
ground-truth initial pose, predicts absolute joint angles for an initialization
prefix, and then integrates predicted angular increments. \emph{Tracking}
causally repeats projected features on the 50-Hz grid, receives the
ground-truth pose at the scoring boundary, and integrates predicted increments
from that state. They are parallel benchmark protocols and are reported
separately~\citep{salter2024emg2pose}.

Frozen transfer fixes the hidden-state backbone and trains only the feature
projection and recurrent decoder, probing representation quality
~\citep{yang2021superb}. Full-backbone adaptation warm-starts the frozen pose
model and additionally trains the EMG-GPT encoder. NeuroRVQ, its codebooks, and the bypassed categorical head remain fixed in every evaluation protocol.

\section{Evaluation}
\label{sec:evaluation}

We preserve the emg2pose partitions and report Regression separately from
Tracking, which receives a ground-truth boundary pose~\citep{salter2024emg2pose}.
\emph{User} holds out participants, \emph{Stage} holds out kinematic
categories, and \emph{User, Stage} holds out both. 
During evaluation we use $K=4$, 25-Hz representation and full-backbone adaptation (see Appendix \ref{app:implementation}). Tracking adds velocity supervision. 
Selection details are described in
Appendices~\ref{app:pose-protocols}. Appendix~\ref{app:evaluation} reports details evaluation results. 

\begin{figure}[h!]
  \centering
  \includegraphics[width=\linewidth]{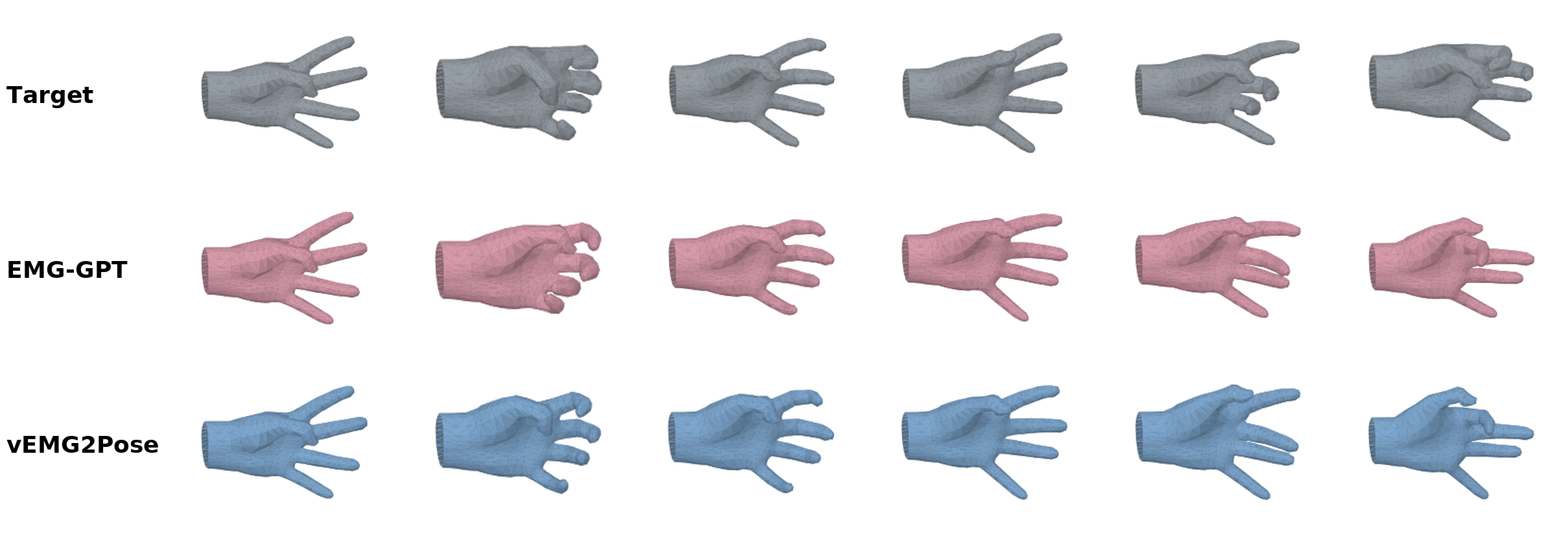}
  \caption{Tracking poses for EMG-GPT and vEMG2Pose. 
  Top: motion capture; middle: EMG-GPT Tracking;
  bottom: vEMG2Pose ~\citep{salter2024emg2pose}. }
  \label{fig:qual-median}
\end{figure}

\begin{table}[h!]
  \centering
  \footnotesize
  \caption{Regression test results. Published baselines are from
  \citet{salter2024emg2pose}; EMG-GPT is a fully adapted model. Angular and landmark values are mean $\pm$ sample SD across users within each generalization condition.}
  \label{tab:main-regression-test}
  \begin{tabular}{@{}llcc@{}}
\toprule
Test set & Model & Angular MAE ($^\circ$) $\downarrow$ & Landmark (mm) $\downarrow$ \\
\midrule
User & SensingDynamics & 15.5 $\pm$ 1.4 & 21.8 $\pm$ 2.1 \\
 & NeuroPose & 13.2 $\pm$ 1.1 & 17.5 $\pm$ 1.3 \\
 & vEMG2Pose & \textbf{12.2 $\pm$ 1.3} & \textbf{15.8 $\pm$ 1.9} \\
 & EMG-GPT & 12.8 $\pm$ 1.1 & 16.3 $\pm$ 1.3 \\
\addlinespace[1pt]
\hline
Stage & SensingDynamics & 18.8 $\pm$ 1.6 & 26.6 $\pm$ 2.0 \\
 & NeuroPose & 17.2 $\pm$ 1.7 & 24.0 $\pm$ 2.1 \\
 & vEMG2Pose & \textbf{15.2 $\pm$ 1.6} & \textbf{20.4 $\pm$ 2.2} \\
 & EMG-GPT & 16.3 $\pm$ 1.6 & 22.0 $\pm$ 1.9 \\
\addlinespace[1pt]
\hline
User, Stage & SensingDynamics & 18.7 $\pm$ 1.6 & 27.2 $\pm$ 2.0 \\
 & NeuroPose & 17.5 $\pm$ 1.5 & 24.9 $\pm$ 1.7 \\
 & vEMG2Pose & \textbf{15.8 $\pm$ 1.4} & \textbf{21.6 $\pm$ 2.0} \\
 & EMG-GPT & 16.6 $\pm$ 1.3 & 22.7 $\pm$ 1.4 \\
\bottomrule
\end{tabular}

\end{table}

Across all three conditions, Regression has lower angular and landmark errors
than SensingDynamics and NeuroPose, while vEMG2Pose remains marginally stronger. Tracking displays good results and remains only marginally behind vEMG2Pose (see Appendix~\ref{app:evaluation}). Because the external results differ in frontend,
augmentation, temporal alignment, and scoring grid, these tables support
descriptive comparison rather than a significance or benchmark-superiority
claim. We refer the reader to Appendix~\ref{app:evaluation} for a matching evaluation protocol.

\newpage
\section{Conclusions}
\label{sec:conclusions}

In this work, we introduce EMG-GPT, an transformer-based model that combines frozen NeuroRVQ geometry, bidirectional
within-frame integration, causal temporal modeling and depth-autoregressive future-code prediction. A series of experiments support transfer from EMG-only pretraining. Future work will investigate joint tokenizer–predictor adaptation, motion-preserving pretraining objectives, and causal streaming inference. More broadly, scaling pretraining across users, devices, and datasets could support a transferable sEMG foundation model for pose estimation, gesture recognition, and human–machine interaction.



\clearpage
\appendix
\section{Frozen NeuroRVQ EMG Tokenization Interface}
\label{app:rvq}

This section describes the NeuroRVQ-EMG tokenizer formulation ~\citep{barmpas2025neurorvq}. It provides the residual-quantization formulation and codebook notation required by our EMG-GPT as described in Section \ref{sec:method}.

\subsection{Codebook assignment and residual recurrence}

For a fixed frame $t$, electrode $c$ and branch $b$, the transformer encoder in NeuroRVQ-EMG tokenizer
produces a latent vector $\mathbf{z}\in\mathbb{R}^{d_e}$. We suppress these indices below, writing
$q_k=q_{t,c,b,k}$ and $\mathbf{E}_k=\mathbf{E}_{b,k}$. Let
$\mathbf{E}_k=\{\mathbf{e}_{k,j}\}_{j=0}^{M-1}$ denote the codebook at RVQ
depth $k$ with $M$ tokens. The location of the selected token in $\mathbf{E}_k$ is denoted as $q_k\in\{0,\ldots,M-1\}$.
With
$\mathbf{r}^{(0)}=\mathbf{z}$ and a numerical stabilizer $\varepsilon>0$, the tokenizer applies:
\begin{align}
  \overline{\mathbf{r}}^{(k-1)}
  &=\frac{\mathbf{r}^{(k-1)}}
  {\max(\lVert\mathbf{r}^{(k-1)}\rVert_2,\varepsilon)},
  \label{eq:app-rvq-normalize}\\
  q_k
  &=\arg\min_{j\in\{0,\ldots,M-1\}}
  \left\lVert\overline{\mathbf{r}}^{(k-1)}-\mathbf{e}_{k,j}\right\rVert_2^2,
  \label{eq:app-rvq-index}\\
  \mathbf{v}^{(k)}&=\mathbf{e}_{k,q_k},
  \qquad
  \mathbf{r}^{(k)}=\mathbf{r}^{(k-1)}-\mathbf{v}^{(k)}.
  \label{eq:app-rvq-update}
\end{align}

The reconstruction from the first $K$ depths is defined as:
\begin{equation}
  \widehat{\mathbf{z}}^{(K)}
  =\sum_{k=1}^{K}\mathbf{v}^{(k)}
  \label{eq:app-rvq-prefix}
\end{equation}

\subsection{Structured EMG output}

NeuroRVQ-EMG tokenizer contains $B=4$ frequency branches. Each branch has a separate
$K_{\max}=16$-level RVQ stack and every codebook contains $M=8192$ token vectors of dimension $d_e=128$. The same branch--depth codebook is shared across
electrodes, but codebooks are not shared across branches or depths. 
NeuroRVQ-EMG is pre-trained on $C=16$ global electrodes.
At frame $t$, $q_{t,c,b,k}\in\{0,\ldots,M-1\}$ is the code index at electrode
$c$, branch $b$, and depth $k$. The prefix
$\mathbf{Q}_t^{(K)}\in\{0,\ldots,M-1\}^{C\times B\times K}$ collects the first
$K$ depths, with reconstructed vector $\widehat{\mathbf{z}}_{t,c,b}^{(K)}$:
\begin{equation}
  \mathbf{Q}_t^{(K)}
  =(q_{t,c,b,k})_{c=1,b=1,k=1}^{C,B,K},
  \qquad
  \widehat{\mathbf{z}}_{t,c,b}^{(K)}
  =\sum_{k=1}^{K}\mathbf{E}_{b,k}[q_{t,c,b,k}].
  \label{eq:app-emg-interface}
\end{equation}



\subsection{Identifiers and frozen geometry}

Code indices are table addresses. For any permutation $\pi_{b,k}$, relabeling
$q'_{t,c,b,k}=\pi_{b,k}(q_{t,c,b,k})$ and permuting the rows of
$\mathbf{E}_{b,k}$ by the same map leaves every selected vector and therefore
Eq.~\eqref{eq:app-emg-interface} unchanged. The relevant geometry is carried by the frozen
vectors in the matching $(b,k)$ codebook. 

\section{EMG-GPT Predictive Interface}
\label{app:emg-gpt-interface}

This section specifies the tensor operations summarized in
Section~\ref{sec:method}. For a minibatch of $N$ windows, the stored input has
shape $[N,L,C,B,K]$, and frozen codebooks have shape $[B,K,M,d_e]$.
We use $d_{\mathrm{model}}$ for the EMG-GPT hidden dimension. 

\subsection{Frame and temporal backbone}

Frozen dequantization first yields:
\begin{equation}
  \widehat{\mathbf z}^{(K)}_{n,t,c,b}
  =\sum_{k=1}^{K}\mathbf E_{b,k}[q_{n,t,c,b,k}]
  \in\mathbb R^{d_e}
  \label{eq:app-gpt-rvq-sum}
\end{equation}
The selected RVQ-sum input path applies LayerNorm before a bias-free linear
projection, then adds channel and branch embeddings:
\begin{equation}
  \mathbf x_{n,t,c,b}
  =\mathbf W_{\mathrm{in}}\operatorname{LN}_{\mathrm{in}}
  (\widehat{\mathbf z}^{(K)}_{n,t,c,b})
  +\mathbf e^{\mathrm{ch}}_c+\mathbf e^{\mathrm{br}}_b
  \label{eq:app-gpt-input-map}
\end{equation}
Reshaping contiguous $[C,B]$ axes produces the channel-major, branch-minor
sequence:
\begin{equation}
  [\mathbf x_{1,1},\ldots,\mathbf x_{1,B},
  \mathbf x_{2,1},\ldots,\mathbf x_{C,B}]
  \label{eq:app-gpt-flattening}
\end{equation}
A learned $\mathrm{[FRAME]}$ token is prepended independently at each time step.
The same bidirectional transformer is applied to all frames, and its
final-layer-normalized $\mathrm{[FRAME]}$ output is $\mathbf f_{n,t}$.

Absolute time embeddings are then added and dropout is applied:
\begin{equation}
  \mathbf U=\operatorname{Drop}
  ([\mathbf f_{n,t}+\mathbf e^{\mathrm{time}}_t]_{n,t})
  \in\mathbb R^{N\times L\times d_{\mathrm{model}}}.
  \label{eq:app-gpt-temporal-input}
\end{equation}
The selected temporal backbone applies causal transformer blocks followed by a
final LayerNorm:
\begin{equation}
  \mathbf H=\operatorname{LN}_{\mathrm{out}}
  F_{\mathrm{GPT}}(\mathbf U)
  =[\mathbf h_1,\ldots,\mathbf h_L]
  \in\mathbb R^{N\times L\times d_{\mathrm{model}}}
  \label{eq:app-gpt-hidden}
\end{equation}
Consequently, $\mathbf h_t$ depends only on token frames $1{:}t$. The dataset
uses complete contiguous windows and pairs
$\mathbf Q_{1:L}$ with $\mathbf Q_{1+\Delta:L+\Delta}$. Discontinuities split
sequences rather than introducing padding. If the token-frame spacing is
$\delta t$, the physical forecast horizon is $\Delta\delta t$.

\subsection{Depth-autoregressive codebook head}

At depth $k$, the head creates one query for every $(c,b)$ position. Before the
decoder blocks, its implemented query is:
\begin{equation}
  \mathbf a_{t,c,b,k}
  =P_h(\mathbf h_t)+\mathbf e^{\mathrm{out,ch}}_c
  +\mathbf e^{\mathrm{out,br}}_b+\mathbf e^{\mathrm{depth}}_k
  +\mathbb I[k>1]\,\mathbf W_{<k}\operatorname{LN}_{<k}
  (\widehat{\mathbf z}^{(<k)}_{t+\Delta,c,b})
  \label{eq:app-gpt-query}
\end{equation}
where $P_h$ is an affine hidden-state projection and $\mathbf W_{<k}$ is a
shared bias-free projection from $d_e$ to $d_{\mathrm{model}}$.
$\operatorname{LN}_{<k}$ is LayerNorm over the $d_e$-dimensional partial
reconstruction $\widehat{\mathbf z}^{(<k)}$, which is defined in
Eq.~\eqref{eq:method-depth-context}. Its value is zero at $k=1$.

Shared decoder blocks first apply bidirectional self-attention across the $CB$
queries at the current depth, then cross-attend each query to the single state
$\mathbf h_t$, and finally apply a multilayer perceptron (MLP). Same-depth queries therefore exchange
hidden context, but no realized same-depth target is supplied. Training uses
ground-truth coarser codes in Eq.~\eqref{eq:app-gpt-query}, and generation replaces
them with previously generated coarser codes.

Let $\mathbf g_{t,c,b,k}$ be the final-layer-normalized query after these
blocks and let $P_q:\mathbb R^{d_{\mathrm{model}}}\rightarrow\mathbb R^{d_e}$
be the shared affine output projection. Logits for code index $j$ are defined as:
\begin{equation}
  \ell_{t,c,b,k,j}
  =s_{b,k}\left\langle
  \frac{P_q(\mathbf g_{t,c,b,k})}
       {\lVert P_q(\mathbf g_{t,c,b,k})\rVert_2},
  \frac{\mathbf E_{b,k}[j]}
       {\lVert\mathbf E_{b,k}[j]\rVert_2}
  \right\rangle,
  \qquad
  s_{b,k}=\min\{\exp(\gamma_{b,k}),s_{\max}\}
  \label{eq:app-gpt-codebook-logits}
\end{equation}
where $\gamma_{b,k}$ is a learned branch--depth log-scale and $s_{\max}$ is a
fixed upper bound on $s_{b,k}$.
Thus each branch and depth is scored only against its matching frozen
codebook. Decoder blocks and $P_q$ are shared across depth. Depth embeddings and
positive logit scales are depth-specific and branch--depth-specific,
respectively. Channel and branch embeddings are
shared over depth. The primary loss averages branch-wise summed
cross-entropies over $NLCB$ targets at each depth and then applies the normalized
depth weights in Eq.~\eqref{eq:method-pretraining-loss}.

\subsection{Effect of retained depth and transfer tensor}

Increasing $K$ adds vectors to Eq.~\eqref{eq:app-gpt-rvq-sum} and ordered target
levels to the head. It does not change the $CB$ intra-frame positions, context
length $L$ or number of transferred temporal states. Transfer consumes exactly
$\mathbf H$ from Eq.~\eqref{eq:app-gpt-hidden}, after the final backbone
LayerNorm and before the categorical head. It, therefore, receives neither code
logits nor predicted future code indices.

\section{Implementation and Pretraining Details}
\label{app:implementation}

Training slices the
first $K$ depths and constructs complete length-$L$ windows paired with targets
$\Delta$ token frames ahead. We use 25-Hz tokens with a 40-ms hop and $\Delta=5$. This configuration predicts 200~ms ahead.
Patch width is recorded separately and is not used to define this
forecast interval. Missing storage blocks split a recording into separate
sequences, so a window never spans a discontinuity. 
Codebooks are loaded as non-trainable runtime buffers with shape
$[B,K,M,d_e]$. 
The input path and categorical head use
the same frozen branch--depth tables, while all frame-encoder, temporal, and
query-projection parameters are learned during pretraining.

\begin{table}[h!]
  \centering
  \small
  \caption{Configuration of the selected 25-Hz, $K=4$ EMG-GPT model.}
  \label{tab:implementation-final}
  \begin{tabular}{@{}p{0.28\linewidth} | p{0.64\linewidth}@{}}
    \toprule
    {\raggedright Field} & {\raggedright Verified value} \\
    \midrule
    Token grid & 25 Hz; 40-ms hop; independently encoded 200-ms patches \\
    RVQ depth & $K=4$ retained depths \\
    Forecast & $L=150$ frames (6.0 s context); $\Delta=5$ frames (200 ms) \\
    Offline preprocessing & Nominal 20--400 Hz third-order Butterworth with
      forward--reverse filtering; polyphase resampling from 2 kHz to 1 kHz \\
    Input representation & Frozen $K=4$ RVQ vectors summed across depth,
      LayerNorm, bias-free projection, learned channel/branch embeddings, CLS pooling \\
    Temporal backbone & $d_{\mathrm{model}}=512$; 8 Transformer layers; 8 heads;
      dropout 0.1; 2-layer within-frame encoder \\
    Output head & 2-layer depth-autoregressive codebook head; uniform depth weighting \\
    Parameters & 40,494,736 trainable parameters (40.5M) \\
    Objective & Hard categorical CE plus logit z-loss with $\lambda_z=10^{-4}$ \\
    Optimizer & AdamW; peak LR $10^{-4}$; $\beta=(0.9,0.95)$; weight decay 0.1 \\
    Batch and warmup & Batch 16; 2 gradient-accumulation steps; effective batch 32;
      2,000-step linear warmup \\
    Schedule & WSD through 200k, with cosine decay over steps 180k--200k to
      $10^{-5}$; constant $10^{-5}$ continuation from 200k to 400k \\
    Precision and hardware & bfloat16 on one NVIDIA GH200 120GB; seed 123 \\
    Exact 400k exposure & 12.8M windows; 1.92B predicted frames;
      491.52B predicted code targets \\
    Validation-selected GPT & Regression: 280k; Tracking: 400k \\
    \bottomrule
  \end{tabular}
\end{table}


\section{Regression and Tracking Protocols}
\label{app:pose-protocols}

This section specifies the emg2pose evaluation protocols. Both tasks
use the final normalized states of a pretrained $K=4$ EMG-GPT. NeuroRVQ and its codebooks remain
fixed. Inputs contain 150 token frames at 25~Hz (6~s), including 25 left-context
frames (1~s). Pose targets are sampled at 50~Hz. Invalid inverse-kinematics rows
are masked from optimization and scoring.

\paragraph{Pose decoders and alignment.}
A learned projection maps each 512-dimensional backbone state to 64 features. Both
tasks then use a two-layer LSTM with hidden width 512 to predict 20 joint
angles (see Figure \ref{fig:method-overview}). \textit{Regression} linearly interpolates token features to 50~Hz with endpoint
alignment. It receives no ground-truth initial pose: the decoder starts from
zeros, predicts absolute pose for 12 steps, and subsequently integrates angular
increments scaled by 0.01. The scored tail contains 250 pose steps (5~s), and
windows advance by 150 token frames (6~s). \textit{Tracking} uses a causal zero-order hold, repeating every 25-Hz feature for two
pose steps. It receives the ground-truth pose at the scoring boundary, 1.04~s
after the token-window start, and then autoregressively integrates increments
scaled by 0.01. Its one-token-frame offset gives 40~ms token-to-pose latency. The
scored tail is again 5~s, while windows advance by 125 token frames (5~s).

\paragraph{Objectives.}
The angular term is masked joint-angle L1 in radians. The fingertip term is the mean distance in millimetres over the five fingertips. Regression uses angular L1 plus $0.01$ fingertip distance. Tracking adds velocity loss with weight 4. Full adaptation trains the RVQ-input mapping and intra-frame encoder, all temporal GPT blocks, temporal and structural embeddings, and final backbone LayerNorm. It uses Adam with an
effective batch of 32, head learning rate $5\times10^{-5}$, backbone learning
rate $5\times10^{-6}$, and a 5k-step cosine schedule.

\section{Transfer Evaluation}
\label{app:evaluation}

\subsection{Tracking Results}

\begin{table}[h!]
  \centering
  \footnotesize
  \caption{Tracking test results, where the boundary pose is given. 
  vEMG2Pose values are from \citep{salter2024emg2pose}.}
  \label{tab:main-tracking-test}
  \begin{tabular}{@{}llcc@{}}
\toprule
Test set & Model & Angular MAE ($^\circ$) $\downarrow$ & Landmark (mm) $\downarrow$ \\
\midrule
User & vEMG2Pose & \textbf{7.7 $\pm$ 1.0} & \textbf{10.3 $\pm$ 1.5} \\
 & EMG-GPT & 8.9 $\pm$ 0.9 & 11.6 $\pm$ 1.2 \\
\addlinespace[1pt]
\hline
Stage & vEMG2Pose & \textbf{11.2 $\pm$ 1.4} & \textbf{15.2 $\pm$ 1.9} \\
 & EMG-GPT & 12.4 $\pm$ 1.4 & 16.9 $\pm$ 1.8 \\
\addlinespace[1pt]
\hline
User, Stage & vEMG2Pose & \textbf{11.0 $\pm$ 1.0} & \textbf{15.4 $\pm$ 1.4} \\
 & EMG-GPT & 12.2 $\pm$ 1.3 & 16.9 $\pm$ 1.4 \\
\bottomrule
\end{tabular}

\end{table}

\subsection{Matching Evaluation Protocol with vEMG2Pose}

Published vEMG2Pose values in the main tables (Tables \ref{tab:main-regression-test} and \ref{tab:main-tracking-test}) are external protocol references. To control for differences in evaluation sampling, we also evaluate the
released vEMG2Pose Regression and Tracking checkpoints, without retraining, at
the canonical test target indices used for EMG-GPT. Within each task, the
ordered target-index hashes, valid-value counts, metric implementation, and
equal-user aggregation are identical between models. Table~\ref{tab:appendix-matched-vemg2pose-test} controls the effective test samples and aggregation locally. This local comparison is not a
reproduction of the published protocol: vEMG2Pose retains its released raw-EMG
frontend and weights, while EMG-GPT uses fixed, independently tokenized 200-ms patches, 25-Hz GPT features, and source-aligned 50-Hz scoring. Zero-phase
filtering also makes the complete token pipeline offline. We therefore treat
both comparisons as descriptive and do not make significance or
benchmark-superiority claims.

\begin{table}[h!]
  \centering
  \footnotesize
  \caption{Locally matched test evaluation of the released vEMG2Pose
  checkpoints and EMG-GPT. Values are mean $\pm$ sample SD across users within
  each generalization condition. Bold indicates the lower mean; rounded ties
  are bolded together. Each method retains its native input frontend.}
  \label{tab:appendix-matched-vemg2pose-test}
  \begin{tabular}{@{}llcc@{}}
\toprule
Test set & Model & Angular MAE ($^\circ$) $\downarrow$ & Landmark (mm) $\downarrow$ \\
\midrule
\multicolumn{4}{@{}l}{\textbf{Regression}} \\
User & vEMG2Pose & 12.80 $\pm$ 1.31 & \textbf{16.29 $\pm$ 1.85} \\
 & EMG-GPT & \textbf{12.80 $\pm$ 1.08} & 16.30 $\pm$ 1.29 \\
\addlinespace[1pt]
\hline
Stage & vEMG2Pose & \textbf{15.40 $\pm$ 1.59} & \textbf{20.43 $\pm$ 2.09} \\
 & EMG-GPT & 16.30 $\pm$ 1.57 & 22.02 $\pm$ 1.92 \\
\addlinespace[1pt]
\hline
User, Stage & vEMG2Pose & \textbf{15.73 $\pm$ 1.38} & \textbf{21.25 $\pm$ 1.79} \\
 & EMG-GPT & 16.64 $\pm$ 1.30 & 22.70 $\pm$ 1.39 \\
\midrule \midrule
\multicolumn{4}{@{}l}{\textbf{Tracking}} \\
User & vEMG2Pose & \textbf{8.57 $\pm$ 0.95} & \textbf{11.11 $\pm$ 1.35} \\
 & EMG-GPT & 8.89 $\pm$ 0.93 & 11.59 $\pm$ 1.24 \\
\addlinespace[1pt]
\hline
Stage & vEMG2Pose & \textbf{11.50 $\pm$ 1.41} & \textbf{15.45 $\pm$ 1.81} \\
 & EMG-GPT & 12.42 $\pm$ 1.43 & 16.88 $\pm$ 1.80 \\
\addlinespace[1pt]
\hline
User, Stage & vEMG2Pose & \textbf{11.33 $\pm$ 0.97} & \textbf{15.64 $\pm$ 1.33} \\
 & EMG-GPT & 12.18 $\pm$ 1.25 & 16.93 $\pm$ 1.36 \\
\bottomrule
\end{tabular}

\end{table}

On held-out users, the matched Regression angular errors are equal after
rounding to two decimals. vEMG2Pose has lower mean error in the remaining
Regression comparisons and in all Tracking comparisons.




\newpage
\subsection{Qualitative Results}

\begin{figure}[h!]
  \centering
  \includegraphics[width=\linewidth]{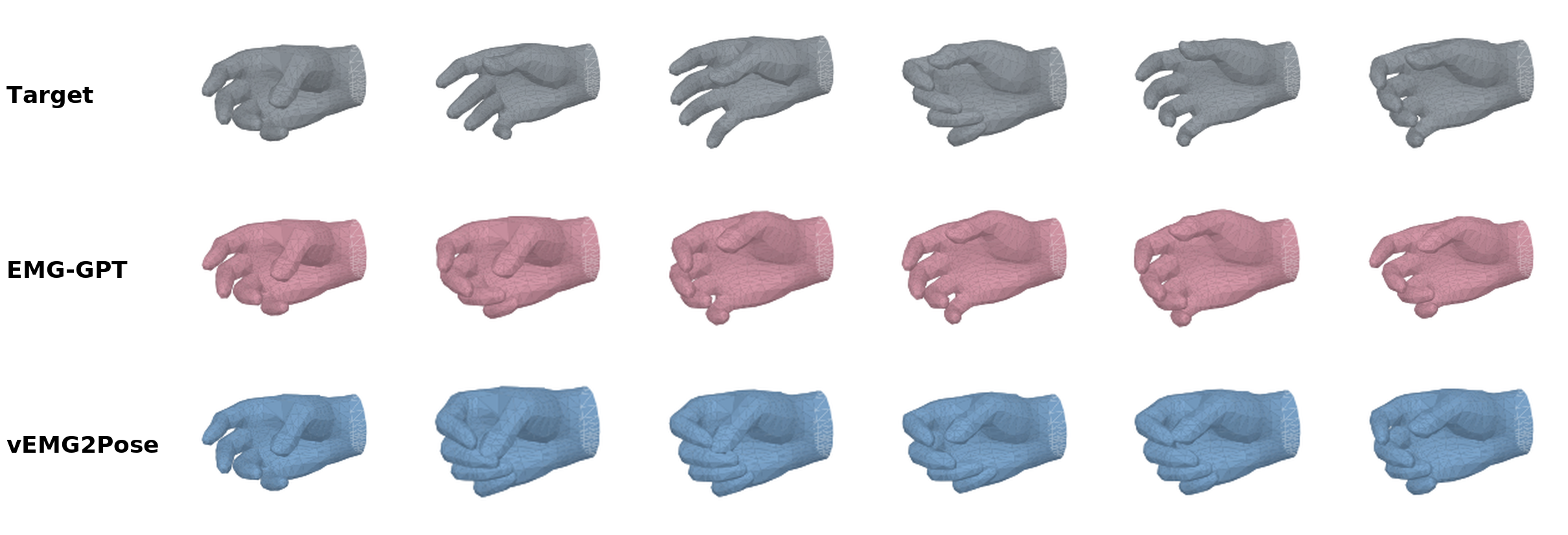}
  \caption{Higher-error held-out user and median held-out stage
  (\textit{FingerFreeform}). Rows and temporal layout follow
  Figure~\ref{fig:qual-median}.}
  \label{fig:qual-user-hard}
\end{figure}

\begin{figure}[h!]
  \centering
  \includegraphics[width=\linewidth]{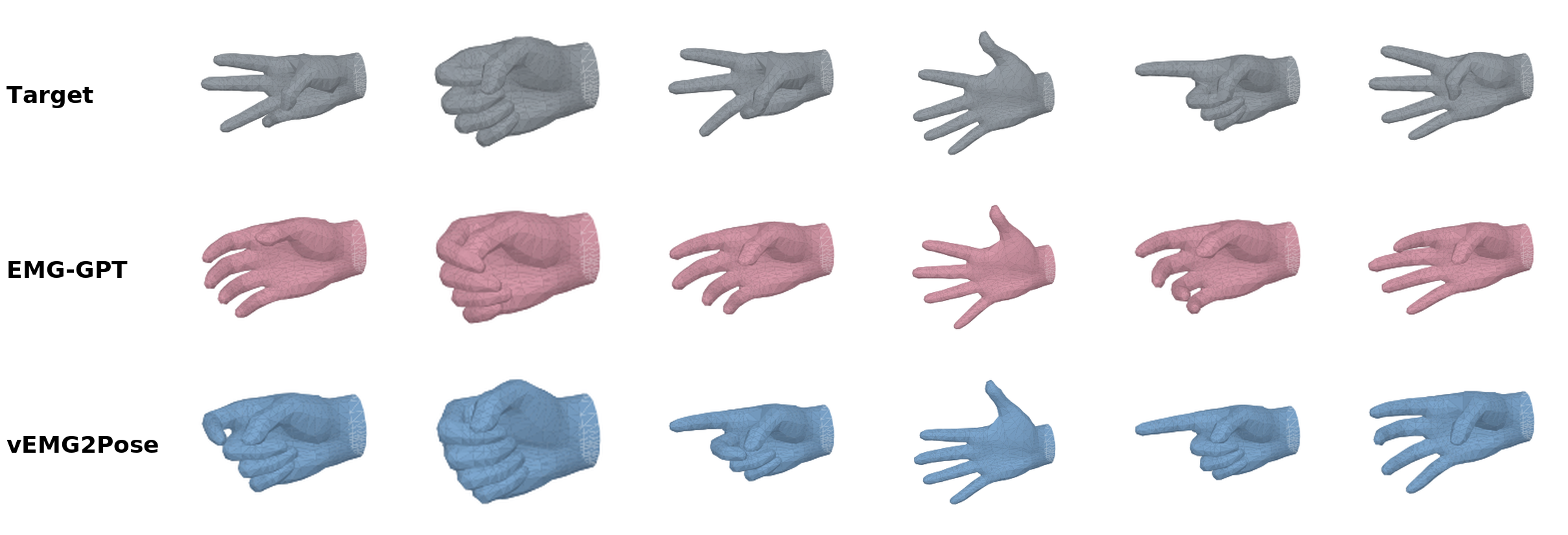}
  \caption{Regression example for a median held-out user and median held-out
  stage. Top: motion-capture target; middle: EMG-GPT Regression; bottom:
  vEMG2Pose Regression. Six poses unroll from left to right over the same
  target-aligned five-second validation clip.}
  \label{fig:qual-reg-median}
\end{figure}

\begin{figure}[h!]
  \centering
  \includegraphics[width=\linewidth]{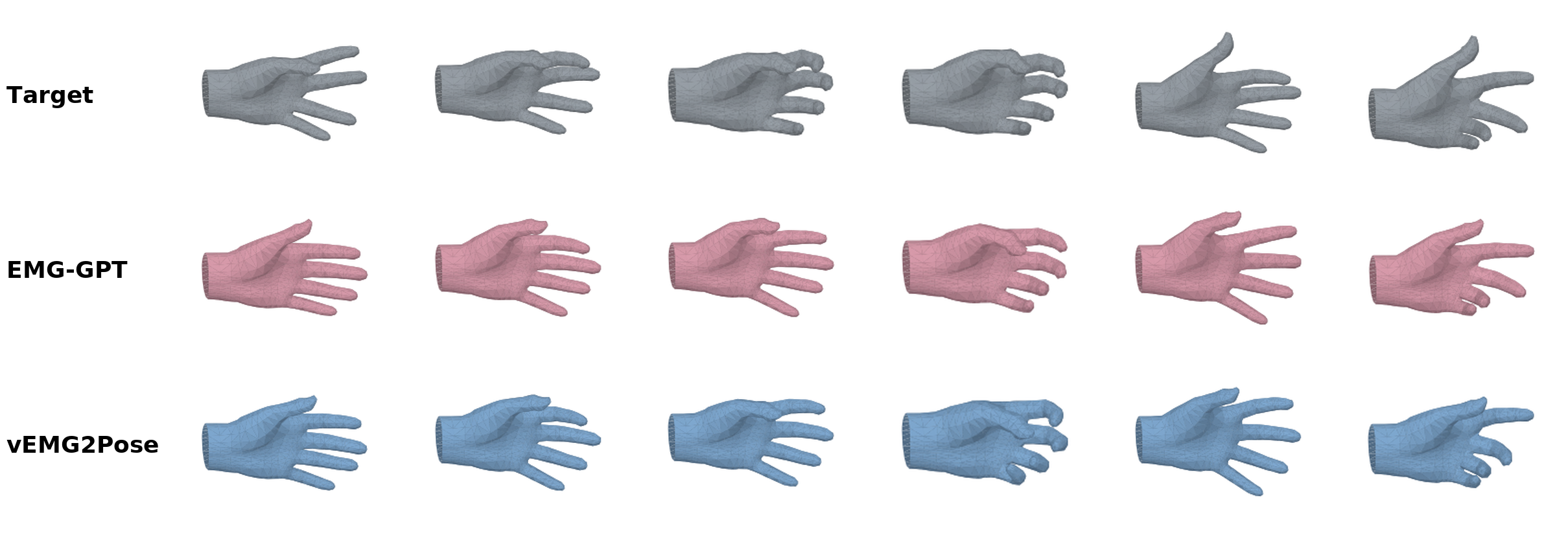}
  \caption{Regression example for a median held-out user and lower-error
  held-out stage. Rows and target-aligned temporal layout follow
  Figure~\ref{fig:qual-reg-median}.}
  \label{fig:qual-reg-easy}
\end{figure}

\begin{figure}[h!]
  \centering
  \includegraphics[width=\linewidth]{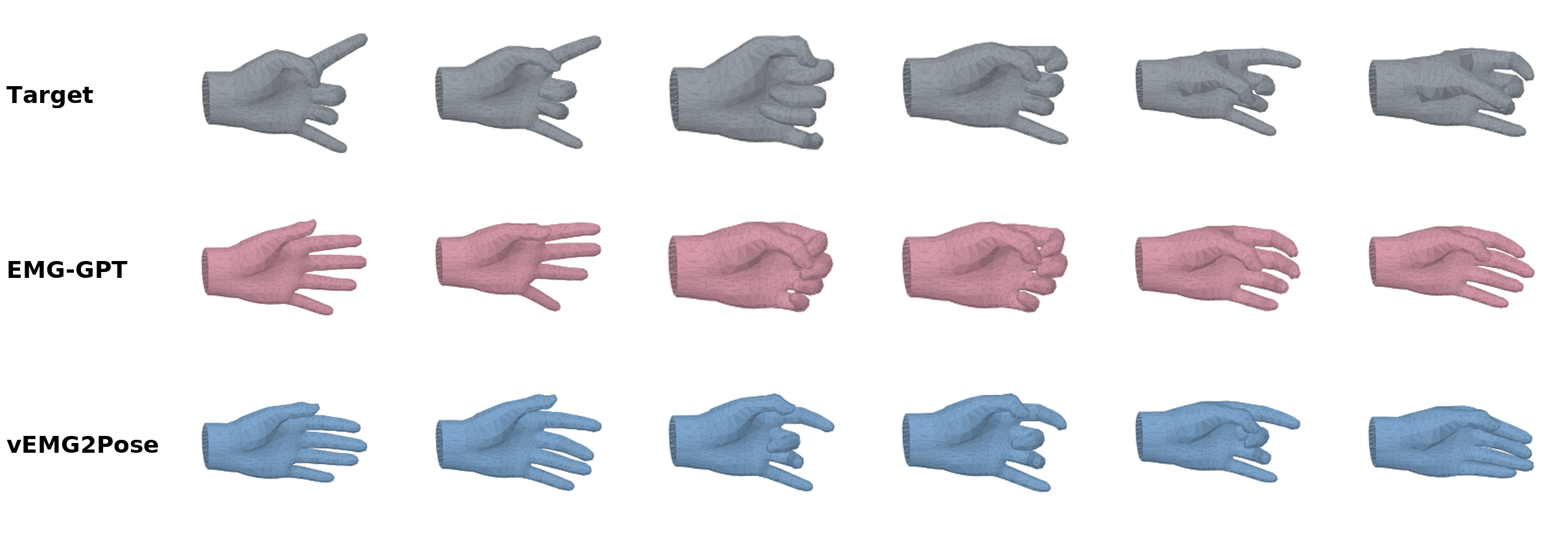}
  \caption{Regression example for a median held-out user and higher-error
  held-out stage. Rows and target-aligned temporal layout follow
  Figure~\ref{fig:qual-reg-median}.}
  \label{fig:qual-reg-hard}
\end{figure}

We select validation clips using pre-defined user and stage error percentiles, then choose windows nearest to the corresponding aggregate error.  Figures~\ref{fig:qual-median}--\ref{fig:qual-user-hard} show the
motion-capture target, the locked 25-Hz EMG-GPT Tracking model, and a locally
evaluated vEMG2Pose reference on the same clips.
Figures~\ref{fig:qual-reg-median}--\ref{fig:qual-reg-hard} apply the same
selection rule to Regression, with target equality verified for each aligned
comparison segment. 

\begin{figure}[h!]
  \centering
  \includegraphics[width=\linewidth]{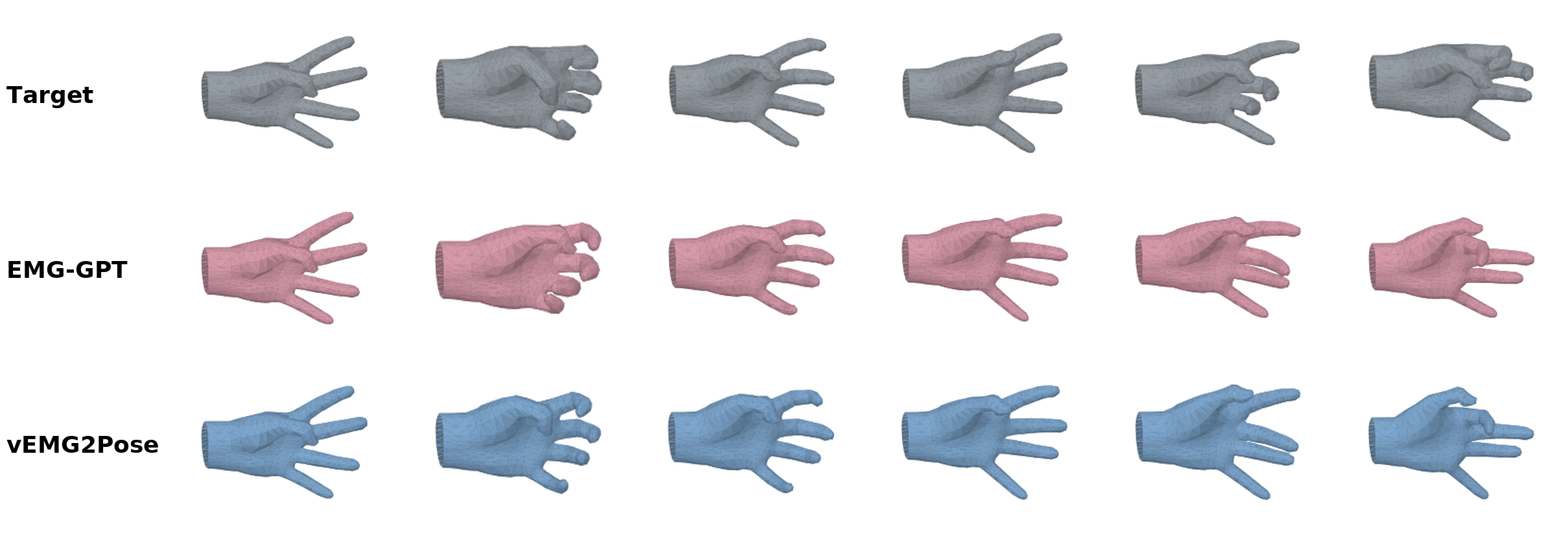}
  \caption{Median held-out user and median held-out stage
  (\textit{FingerFreeform}). Top: motion capture; middle: EMG-GPT Tracking;
  bottom: vEMG2Pose. Six poses unroll from left to right over a five-second
  validation clip.}
  \label{fig:qual-median}
\end{figure}

\begin{figure}[h!]
  \centering
  \includegraphics[width=\linewidth]{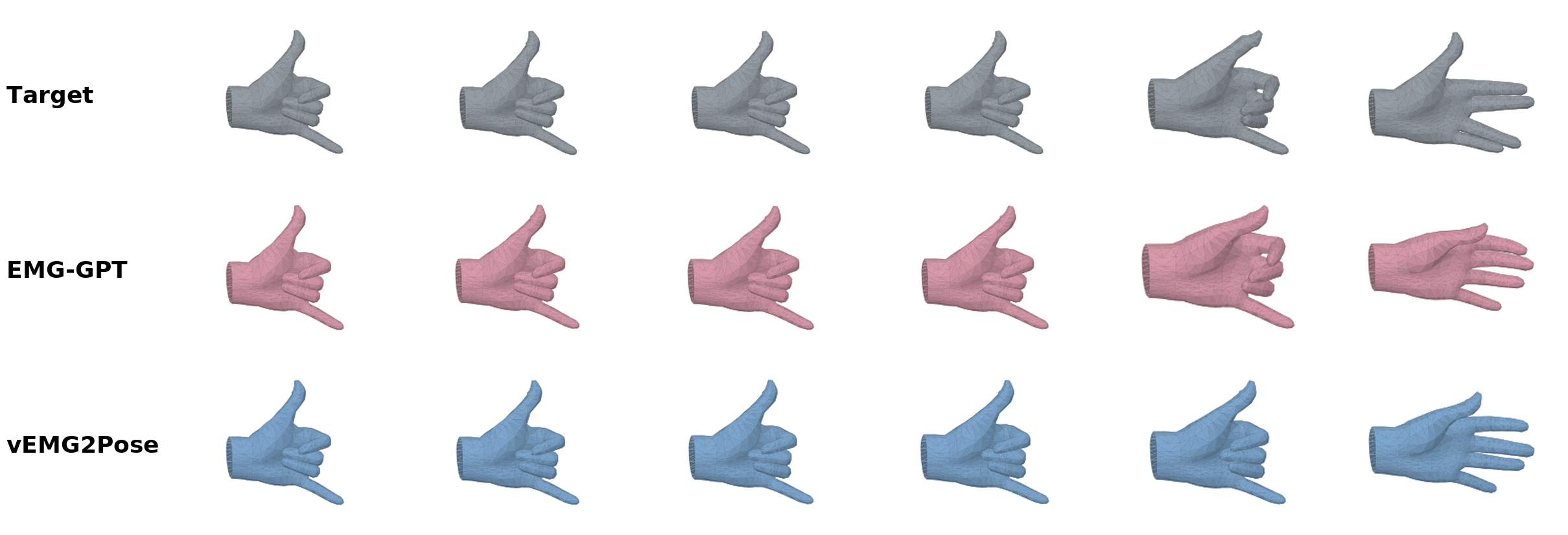}
  \caption{Median held-out user and lower-error held-out stage
  (\textit{ShakaVulcanPeace}). Rows and temporal layout follow
  Figure~\ref{fig:qual-median}.}
  \label{fig:qual-easy}
\end{figure}

\begin{figure}[h!]
  \centering
  \includegraphics[width=\linewidth]{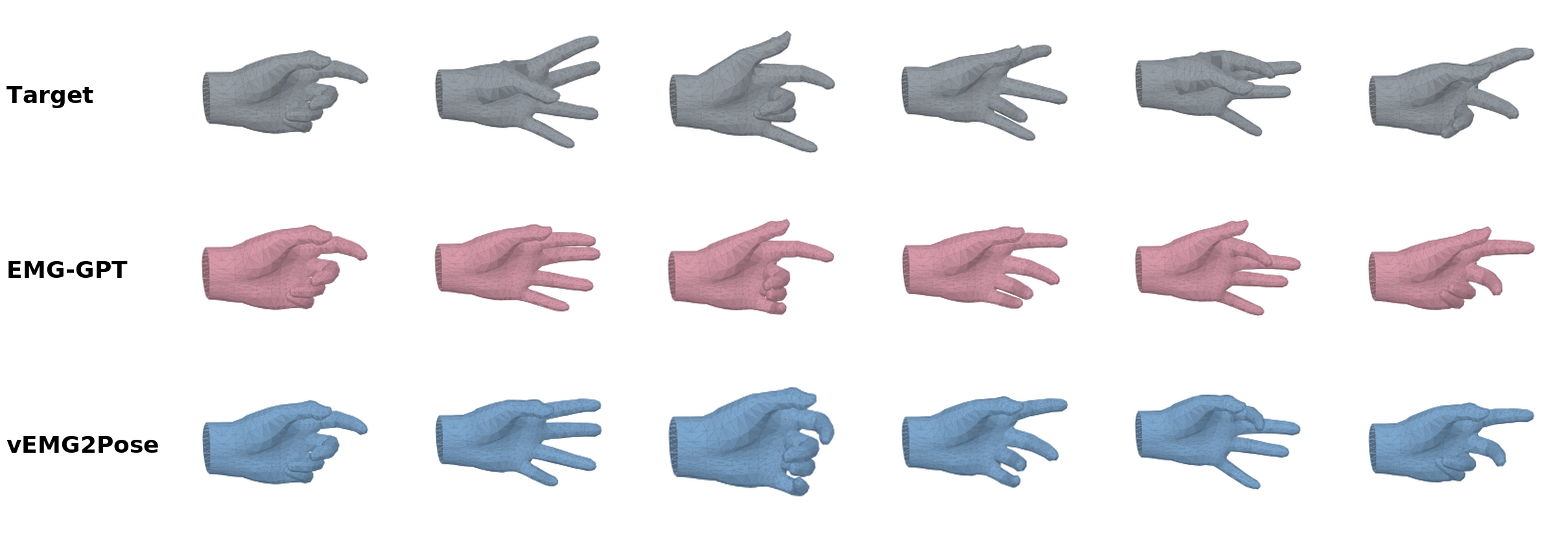}
  \caption{Median held-out user and higher-error held-out stage
  (\textit{CountingUpDownFingerWigglingSpreading}). Rows and temporal layout
  follow Figure~\ref{fig:qual-median}.}
  \label{fig:qual-hard}
\end{figure}

\begin{figure}[h!]
  \centering
  \includegraphics[width=\linewidth]{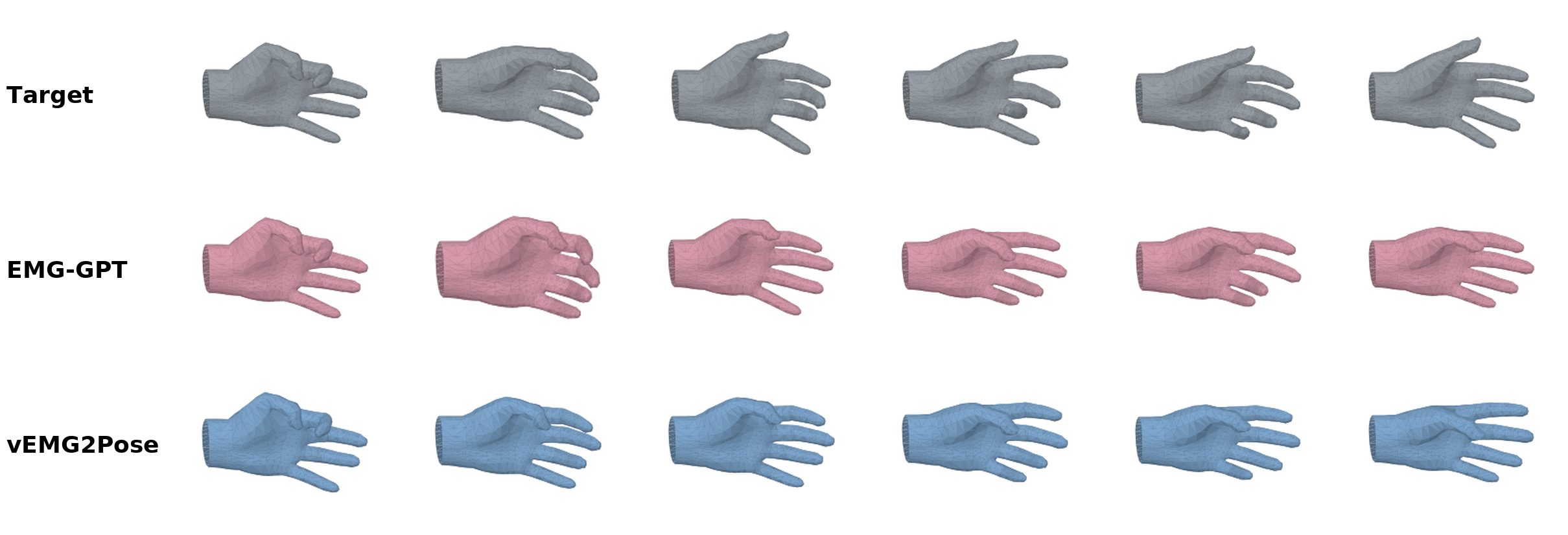}
  \caption{Lower-error held-out user and median held-out stage
  (\textit{HandOverHandCountingUpDownFingerWigglingSpreading}). Rows and
  temporal layout follow Figure~\ref{fig:qual-median}.}
  \label{fig:qual-user-easy}
\end{figure}

\clearpage

\clearpage
\bibliographystyle{plainnat}
\bibliography{references}

\begin{thebibliography}{18}
\providecommand{\natexlab}[1]{#1}
\providecommand{\url}[1]{\texttt{#1}}
\expandafter\ifx\csname urlstyle\endcsname\relax
  \providecommand{\doi}[1]{doi: #1}\else
  \providecommand{\doi}{doi: \begingroup \urlstyle{rm}\Url}\fi

\bibitem[Barmpas et~al.(2026)Barmpas, Lee, Chalatsis, Raftery, Panagakis, Adamos, Laskaris, Koliousis, Farina, and Zafeiriou]{barmpas2025neurorvq}
Konstantinos Barmpas, Na~Lee, Dimitrios Chalatsis, William Raftery, Yannis Panagakis, Dimitrios~A. Adamos, Nikolaos Laskaris, Alexandros Koliousis, Dario Farina, and Stefanos Zafeiriou.
\newblock Neurorvq: Multi-scale biosignal tokenization for generative foundation models, 2026.
\newblock URL \url{https://arxiv.org/abs/2510.13068}.

\bibitem[Cui et~al.(2024)Cui, Jeong, Th{\"o}lke, Medani, Jerbi, Joshi, and Leahy]{cui2024neurogpt}
Wenhui Cui, Woojae Jeong, Philipp Th{\"o}lke, Takfarinas Medani, Karim Jerbi, Anand~A. Joshi, and Richard~M. Leahy.
\newblock {Neuro-GPT}: Towards a foundation model for {EEG}.
\newblock In \emph{IEEE International Symposium on Biomedical Imaging}, pages 1--5, 2024.
\newblock \doi{10.1109/ISBI56570.2024.10635453}.

\bibitem[Jarque-Bou et~al.(2021)Jarque-Bou, Sancho-Bru, and Vergara]{s21093035}
Néstor~J. Jarque-Bou, Joaquín~L. Sancho-Bru, and Margarita Vergara.
\newblock A systematic review of emg applications for the characterization of forearm and hand muscle activity during activities of daily living: Results, challenges, and open issues.
\newblock \emph{Sensors}, 21\penalty0 (9), 2021.
\newblock ISSN 1424-8220.
\newblock \doi{10.3390/s21093035}.
\newblock URL \url{https://www.mdpi.com/1424-8220/21/9/3035}.

\bibitem[Jiang et~al.(2024)Jiang, Zhao, and Lu]{jiang2024labram}
Wei-Bang Jiang, Li-Ming Zhao, and Bao-Liang Lu.
\newblock Large brain model for learning generic representations with tremendous {EEG} data in {BCI}.
\newblock In \emph{International Conference on Learning Representations}, 2024.
\newblock URL \url{https://openreview.net/forum?id=QzTpTRVtrP}.

\bibitem[Lee et~al.(2022)Lee, Kim, Kim, Cho, and Han]{lee2022rqtransformer}
Doyup Lee, Chiheon Kim, Saehoon Kim, Minsu Cho, and Wook-Shin Han.
\newblock Autoregressive image generation using residual quantization.
\newblock In \emph{Proceedings of the IEEE/CVF Conference on Computer Vision and Pattern Recognition}, pages 11523--11532, 2022.
\newblock \doi{10.1109/CVPR52688.2022.01123}.

\bibitem[Liu et~al.(2021)Liu, Zhang, and Gowda]{liu2021neuropose}
Yilin Liu, Shijia Zhang, and Mahanth Gowda.
\newblock {NeuroPose}: 3d hand pose tracking using {EMG} wearables.
\newblock In \emph{Proceedings of the Web Conference 2021}, pages 1471--1482, 2021.
\newblock \doi{10.1145/3442381.3449890}.

\bibitem[Pavlakos et~al.(2023)Pavlakos, Shan, Radosavovic, Kanazawa, Fouhey, and Malik]{pavlakos2023reconstructinghands3dtransformers}
Georgios Pavlakos, Dandan Shan, Ilija Radosavovic, Angjoo Kanazawa, David Fouhey, and Jitendra Malik.
\newblock Reconstructing hands in 3d with transformers, 2023.
\newblock URL \url{https://arxiv.org/abs/2312.05251}.

\bibitem[Potamias et~al.(2025)Potamias, Zhang, Deng, and Zafeiriou]{potamias2025wilorendtoend3dhand}
Rolandos~Alexandros Potamias, Jinglei Zhang, Jiankang Deng, and Stefanos Zafeiriou.
\newblock Wilor: End-to-end 3d hand localization and reconstruction in-the-wild, 2025.
\newblock URL \url{https://arxiv.org/abs/2409.12259}.

\bibitem[Quivira et~al.(2018)Quivira, Koike-Akino, Wang, and Erdogmus]{quivira2018semgpose}
Fernando Quivira, Toshiaki Koike-Akino, Ye~Wang, and Deniz Erdogmus.
\newblock Translating {sEMG} signals to continuous hand poses using recurrent neural networks.
\newblock In \emph{IEEE EMBS International Conference on Biomedical and Health Informatics}, pages 166--169, 2018.
\newblock \doi{10.1109/BHI.2018.8333395}.

\bibitem[Salter et~al.(2024)Salter, Warren, Schlager, Spurr, Han, Bhasin, Cai, Walkington, Bolarinwa, Wang, Danielson, Merel, Pnevmatikakis, and Marshall]{salter2024emg2pose}
Sasha Salter, Richard Warren, Collin Schlager, Adrian Spurr, Shangchen Han, Rohin Bhasin, Yujun Cai, Peter Walkington, Anuoluwapo Bolarinwa, Robert Wang, Nathan Danielson, Josh Merel, Eftychios Pnevmatikakis, and Jesse Marshall.
\newblock {emg2pose}: A large and diverse benchmark for surface electromyographic hand pose estimation.
\newblock In \emph{Advances in Neural Information Processing Systems}, volume~37, 2024.
\newblock \doi{10.52202/079017-1770}.

\bibitem[Si et~al.(2026)Si, Liu, Ai, Xie, Potamias, Zheng, and Su]{si2026anyhand}
Chen Si, Yulin Liu, Bo~Ai, Jianwen Xie, Rolandos~Alexandros Potamias, Chuanxia Zheng, and Hao Su.
\newblock Anyhand: A large-scale synthetic dataset for rgb (-d) hand pose estimation.
\newblock \emph{arXiv preprint arXiv:2603.25726}, 2026.

\bibitem[Simpetru et~al.(2022)Simpetru, Osswald, Braun, Oliveira, Cakici, and Del~Vecchio]{simpetru2022dof}
Raul~C. Simpetru, Marius Osswald, Dominik~I. Braun, Daniela~S. Oliveira, Andre~L. Cakici, and Alessandro Del~Vecchio.
\newblock Accurate continuous prediction of 14 degrees of freedom of the hand from myoelectric signals through convolutional neural networks.
\newblock In \emph{44th Annual International Conference of the IEEE Engineering in Medicine and Biology Society}, pages 702--706, 2022.
\newblock \doi{10.1109/EMBC48229.2022.9870937}.

\bibitem[Tchantchane et~al.(2023)Tchantchane, Zhou, Zhang, and Alici]{Tchantchane}
Rayane Tchantchane, Hao Zhou, Shen Zhang, and Gursel Alici.
\newblock A review of hand gesture recognition systems based on noninvasive wearable sensors.
\newblock \emph{Advanced Intelligent Systems}, 5\penalty0 (10):\penalty0 2300207, 2023.
\newblock \doi{https://doi.org/10.1002/aisy.202300207}.

\bibitem[Vaswani et~al.(2017)Vaswani, Shazeer, Parmar, Uszkoreit, Jones, Gomez, Kaiser, and Polosukhin]{vaswani2017attention}
Ashish Vaswani, Noam Shazeer, Niki Parmar, Jakob Uszkoreit, Llion Jones, Aidan~N. Gomez, {\L}ukasz Kaiser, and Illia Polosukhin.
\newblock Attention is all you need.
\newblock In \emph{Advances in Neural Information Processing Systems}, volume~30, 2017.
\newblock URL \url{https://proceedings.neurips.cc/paper/7181-attention-is-all-you-need}.

\bibitem[Yang et~al.(2023)Yang, Westover, and Sun]{yang2023biot}
Chaoqi Yang, M.~Brandon Westover, and Jimeng Sun.
\newblock {BIOT}: Biosignal transformer for cross-data learning in the wild.
\newblock In \emph{Advances in Neural Information Processing Systems}, volume~36, 2023.
\newblock \doi{10.52202/075280-3420}.

\bibitem[Yang et~al.(2021)Yang, Chi, Chuang, Lai, Lakhotia, Lin, Liu, Shi, Chang, Lin, Huang, Tseng, Lee, Liu, Huang, Dong, Li, Watanabe, Mohamed, and Lee]{yang2021superb}
Shu-wen Yang, Po-Han Chi, Yung-Sung Chuang, Cheng-I~Jeff Lai, Kushal Lakhotia, Yist~Y. Lin, Andy~T. Liu, Jiatong Shi, Xuankai Chang, Guan-Ting Lin, Tzu-Hsien Huang, Wei-Cheng Tseng, Ko-tik Lee, Da-Rong Liu, Zili Huang, Shuyan Dong, Shang-Wen Li, Shinji Watanabe, Abdelrahman Mohamed, and Hung-yi Lee.
\newblock {SUPERB}: Speech processing universal {PERformance} benchmark.
\newblock In \emph{Interspeech 2021}, pages 1194--1198, 2021.
\newblock \doi{10.21437/Interspeech.2021-1775}.

\bibitem[Yue et~al.(2024)Yue, Gao, Xue, Tang, Guo, Jiang, and Liu]{yue2025braingpt}
Tongtian Yue, Xuange Gao, Shuning Xue, Yepeng Tang, Longteng Guo, Jie Jiang, and Jing Liu.
\newblock {BrainGPT}: Unleashing the potential of {EEG} generalist foundation model by autoregressive pre-training.
\newblock \emph{arXiv preprint arXiv:2410.19779}, 2024.
\newblock \doi{10.48550/arXiv.2410.19779}.
\newblock Version 2, preprint.

\bibitem[Zeghidour et~al.(2022)Zeghidour, Luebs, Omran, Skoglund, and Tagliasacchi]{zeghidour2022soundstream}
Neil Zeghidour, Alejandro Luebs, Ahmed Omran, Jan Skoglund, and Marco Tagliasacchi.
\newblock {SoundStream}: An end-to-end neural audio codec.
\newblock \emph{IEEE/ACM Transactions on Audio, Speech, and Language Processing}, 30:\penalty0 495--507, 2022.
\newblock \doi{10.1109/TASLP.2021.3129994}.

\end{thebibliography}


\end{document}